\documentclass[a4paper,twocolumn]{article}
\usepackage{listings,float,graphicx,dblfloatfix,amsmath}
\usepackage[columnsep = 0.5cm]{geometry}
\usepackage[round,colon,authoryear]{natbib}
\usepackage{charter,fancyhdr,placeins}
\usepackage[footnotesize]{caption}

\fancypagestyle{plain}{%
        \fancyhf{}
        \fancyfoot[C]{\thepage}
        
        }

\title{\rule{\linewidth}{0.5mm} \\ \vspace{5pt} A Biologically Realistic Model of Saccadic Eye Control with Probabilistic Population Codes \vspace{-4pt} \\ \rule{\linewidth}{0.5mm}}
\author{ \textbf{Sacha Sokoloski} \\ BCCN Berlin \\ sacha@cs.toronto.edu \vspace{5pt} }
\date{}

\begin{document}

\twocolumn[
    \begin{@twocolumnfalse}
        \maketitle
        \thispagestyle{plain}
    \end{@twocolumnfalse} ]

\section{Introduction}

In this paper I will present a new model of eye control developed within the Probabilistic Population Code (PPC) framework. The posterior parietal cortex is believed to direct eye movements, especially in regards to target tracking tasks, and a number of debates exist over the precise nature of the computations performed by the parietal cortex, with each side supported by different sets of biological evidence. In this paper I will present my model which navigates a course between some of these debates, towards the end of presenting a model which can explain some of the competing interpretations among the data sets. In particular, rather than assuming that proprioception or efference copies form the key source of information for computing eye position information, I use a biological plausible implementation of a Kalman filter to optimally combine the two signals, and a simple gain control mechanism in order to accommodate the latency of the proprioceptive signal. Fitting within the Bayesian brain hypothesis, the result is a Bayes optimal solution to the eye control problem, with a range of data supporting claims of biological plausibility.

\section{Computation in the Parietal Cortex}
\label{sec:parietalcortex}

Contemporary research has shown that the posterior parietal cortex is involved in target reaching. In order to accomplish this, the parietal cortex must integrate multi sensory information to derive target and body positions, coordinate eye movements to track targets, and finally coordinate motor activity to reach toward targets.

The computational characteristics of the parietal cortex are well characterized and the subject of ongoing research. Neurons in the parietal cortex exhibit spatial receptive fields selective for target and body positions in various coordinate systems. Retinal coordinates have been observed to be especially common in the parietal cortex, and eye position information is thought to be required to transform variables of interest into and out of a standard retinal coordinate system \citep{cohen_common_2002}. The source and nature of this eye position information is subject to debate, with some arguing for static eye position and others for differential eye position information respectively as the key factor in performing these transformations \citep{blohm_decoding_2009,xu_time_2011,wang_eye_2012}.

In general, population activity can be partially characterized by its gain - a scaling up or down of the total output of the network. Gain normally indicates the precision of an encoding in a population, and maintaining gain at various stages of neural processing involves respecting the precision information of upstream populations. The nonlinear operation known as \textit{divisive normalization} - which generally involves dividing a product of gains by their sum - has been proven to possess certain powerful optimality properties, and has been found to be frequently applied in the low level sensory processes of the brain \citep{schwartz_natural_2001}. 

Parietal neurons also exhibit a phenomenon known as gain field modulation in which gain encodes some other variable of interest such as eye position rather than the usual precision information \citep{blohm_fields_2009}. This mechanism of gain field modulation may also underlie powerful and efficient means of computing coordinate transformations \citep{pouget_computational_2000,chang_using_2009}, indicating that more than simply being a novel way of encoding information, gain field modulation is a critical aspect of certain neural computations.

\section{Probabilistic Population Codes}

Within the probabilistic population code framework, a population of neurons is interpreted as encoding a posterior distribution over a stimulus through the stochastic response of that population to the stimulus. In this part of the review I will introduce the basic case of linear, 1-dimensional PPCs. This technical review has been drawn from the articles \cite{pouget_computational_2000,ma_bayesian_2006,beck_marginalization_2011}, and the simulations and plots were developed by myself. This review is intended only to provide sufficient detail to follow the model presented later. Readers wishing to simulate these models themselves are advised to consult the aforementioned papers.

In general, computation in a PPC involves sampling a population's response $\mathbf{r}$ to a stimulus $s$, the distribution of which is assumed to follow some likelihood distribution $p(\mathbf{r} \mid s)$. A Linear PPC is one in which this likelihood distribution is limited to be one from the exponential family with linear sufficient statistics - that is, an exponential family distribution where the kernel $\mathbf{h}$ is a linear function of $s$ (\ref{eq:likelihood}). This equation for the likelihood also contains a multiplier $\phi(\mathbf{r})$ which corresponds to the gain on the encoding, through which we can incorporate precision information or gain field modulation.

\begin{equation}
    p(\mathbf{r} \mid s) = \phi(\mathbf{r})e^{h(s)*\mathbf{r}}
    \label{eq:likelihood}
\end{equation}

Given an $\mathbf{r}$ sampled from this likelihood distribution and a prior over the stimulus $\alpha(s)$, we may decode a posterior distribution $p(s \mid \mathbf{r})$ over $s$ up to a normalizing term $Z$ (\ref{eq:posterior}). With this distribution we can estimate a value for the stimulus encoded by a particular population's response.

\begin{equation}
    p(s \mid \mathbf{r}) = \frac{e^{h(s)*\mathbf{r} + \alpha(s)}}{Z}
    \label{eq:posterior}
\end{equation}

\begin{figure}
    \centering
    \includegraphics[width=0.475\textwidth,clip=true]{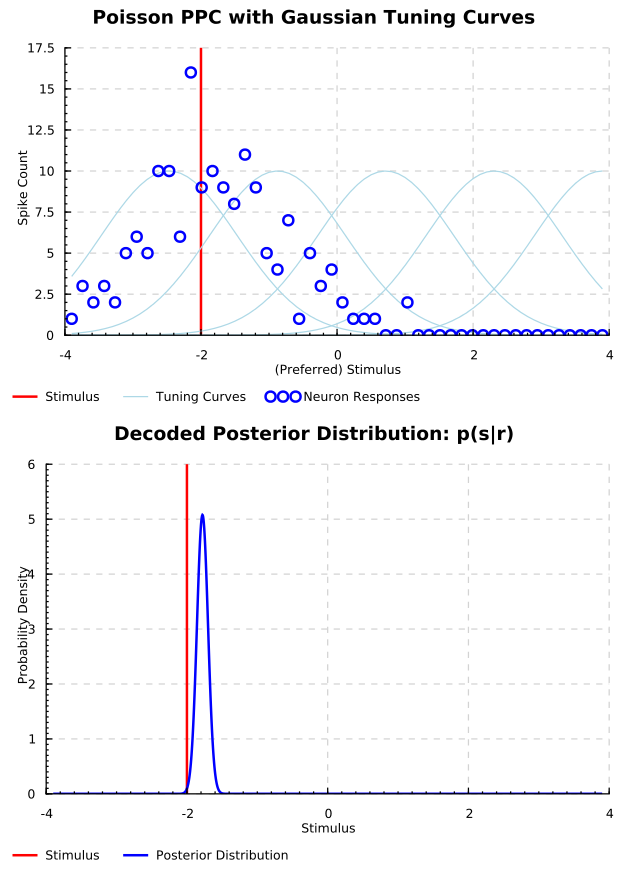}
    \caption{\footnotesize{\textbf{Static Poisson PPC Response}: Top) A sample response of a PPC from the likelihood $p(\mathbf{r} \mid s)$ given the stimulus. Individual neuron responses are arranged over the x axis in accordance with their respective preferred stimuli. The Gaussian tuning curves are also illustrated. Bottom) The decoded Gaussian posterior distribution over the stimulus given the response from the above plot. }}
    \label{fig:static1}
\end{figure}

Although research has demonstrated that neural activity is not always well modelled by Poisson processes with stimulus dependant rates, Poisson models remain useful and essential as first approximation likelihood models. For the simplicity of the mathematics, Gaussian posteriors also tend to be easier to work with and are sufficient for many purposes. If we assume Gaussian tuning curves around some preferred stimulus as defining the rate of our Poisson likelihood distributions $p(\mathbf{r} \mid s)$, then the posterior distribution $p(s \mid \mathbf{r})$ takes a Gaussian form. For the rest of this review, we will assume Poisson distributed likelihoods and Gaussian distributed posteriors. For the case of the static encoding of stimuli, this basic formulation is illustrated in figure~\ref{fig:static1}.

\subsection{Neural Circuits}

Deterministic computation typically involves algorithms which compute the output of a function $y = f(\mathbf{x})$. More generally, however, it may be that the dependence of an output $y$ on an input $x$ is probabilistic rather than deterministic, and that what we therefore wish to compute as output is a posterior distribution $p(y \mid \mathbf{x})$ rather than a single value. More general still, instead of being given true values of the input $\mathbf{x}$ we might be given only distributions $p(\mathbf{x})$ over these true values. In this case we must marginalize over these input distributions in order to calculate an output distribution $p(y)$ (\ref{eq:marginal}). This marginalization will be referred to as a \textit{probabilistic computation}.

\begin{equation}
    p(y) = \int p(y \mid \mathbf{x}) p(\mathbf{x}) d\mathbf{x}
    \label{eq:marginal}
\end{equation}

The PPC framework is general enough to model \textit{neural circuits} - neural populations which implement probabilistic computations. Neural implementations of probabilistic computations, however, are stochastic and do not explicitly encode probability distributions. This means that given a PPC and a sample encoding $\mathbf{r}_{in}$ of some distribution over $\mathbf{x}$, the probabilistic computation implemented by the given PPC is computed implicitly over the given encoding to produce a sample encoding $\mathbf{r}_{out}$ of this posterior distribution $p(y \mid \mathbf{x})$. Moreover, since encodings are sampled from likelihood distributions, there is variability in the encoding of the posterior distributions over $y$.

In general, solving marginalization equations involves solving high dimensional integrals, which is rarely a trivial task. Nevertheless, within the PPC framework Bayes optimal likelihood distributions $p(\mathbf{r}_{out} \mid \mathbf{r}_{in})$ over encodings have been derived for certain interesting cases, providing a basic set of neural circuits which can be composed to model more powerful computations.

 The gain of a neural population is a simple multiplier on population activity, and a key aspect of marginalization in neural circuits involves maintaining appropriate gain in the output encoding. Since larger responses improve the signal to noise ratio, increasing the gain has the effect of increasing the precision of the encoded posterior. In parallel with the biological research mentioned in section~\ref{sec:parietalcortex}, it can be shown that certain optimal computations within the PPC framework implicitly perform divisive normalization, thus supporting the biological plausibility of the PPC approach.

Nevertheless, as also indicated in section~\ref{sec:parietalcortex}, gain need not encode precision alone, and so optimal precision weighting need not be the goal in computing a response to input activities. In this case other variables of interest may be encoded in the sampled responses from equation~\ref{eq:likelihood} via the term $\phi(\mathbf{r})$, and downstream populations may later decode and use the gain modulating signal as required. Deciding what the gain of a PPC encodes is a critical part of applying PPCs, and requires careful consideration of the system being modelled in order to be made correctly.

\subsection{Linear Coordinate Transformations}

Suppose that we're given the position in head centred coordinates of a target $x_{H}$ and 'the eye' $e_{H}$ encoded as appropriate PPC likelihood samples $\tilde{\mathbf{x}}_{H}$ and $\tilde{\mathbf{e}}_{H}$, and we wish to encode the position of the target in eye centred (retinal) coordinates $x_{R}$ as a new response $\tilde{\mathbf{x}}_{R}$. We must then solve the probabilistic computation $p(\tilde{\mathbf{x}}_{R} \mid \tilde{\mathbf{x}}_{H},\tilde{\mathbf{e}}_{H})$ where $x_{R} = x_{H} + e_{H}$. In order to optimize the performance of the computation, the computation must minimize the error in an optimality equation over the posteriors (\ref{eq:optimal}).

\begin{equation}
    p(x_{R} \mid \tilde{\mathbf{x}}_{R}) = p(x_{R} \mid \tilde{\mathbf{x}}_{H},\tilde{\mathbf{e}}_{H})
    \label{eq:optimal}
\end{equation}

\begin{figure}
    \centering
    \includegraphics[width=0.475\textwidth,clip=true]{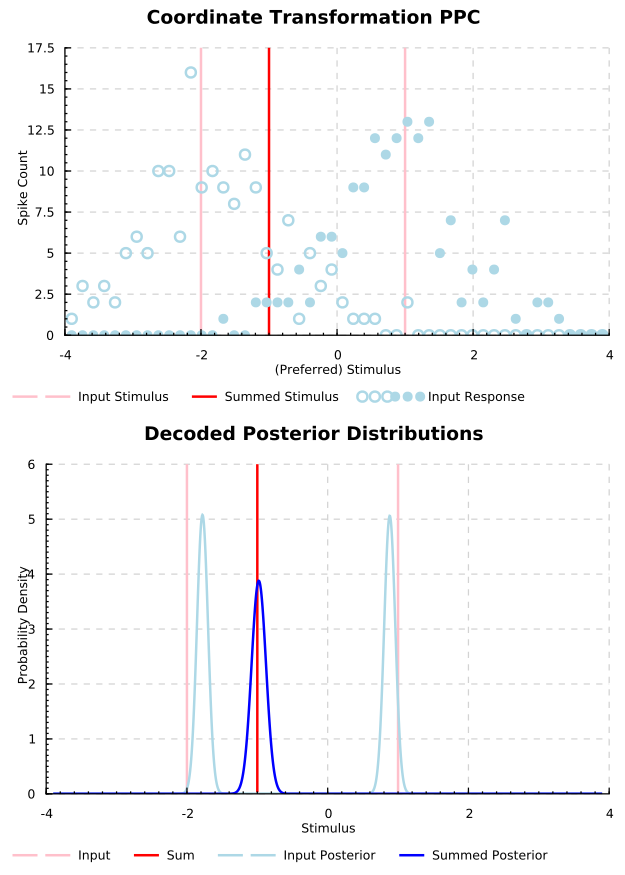}
    \caption{\footnotesize{\textbf{Linear Transformation PPC}: Top) Sample responses from two different populations responding to two distinct stimuli. The summed stimulus is indicated by the darker red. Bottom) The decoded posteriors from the input populations, and the decoded posterior of the linear coordinate transformation PPC which attempts to calculate the sum of the two input populations. The gain (height) of the summing PPC posterior is based on those of the input PPCs via divisive normalization.}}
    \label{fig:static2}
\end{figure}

This Bayesian optimality constraint equates the amount of information we have about $x_{R}$ before and after the computation, thus ensuring that a minimum of information is lost in performing the computation. The results of simulating a computation satisfying the aforementioned conditions are illustrated in figure~\ref{fig:static2}. The gain $g_{3}$ of the optimal output encoding expressed as a function of the input gains $g_{1}$ and $g_{2}$ is given in equation~\ref{eq:divtransform}. This is the expression for divisive normalization with two inputs.

\begin{equation}
    g_{3} = \frac{g_{1} g_{2}}{g_{1} + g_{2}}
    \label{eq:divtransform}
\end{equation}

\subsection{Kalman Filters}
\label{sec:kalman}

\begin{figure*}
    \centering
    \includegraphics[width=0.95\textwidth,clip=true]{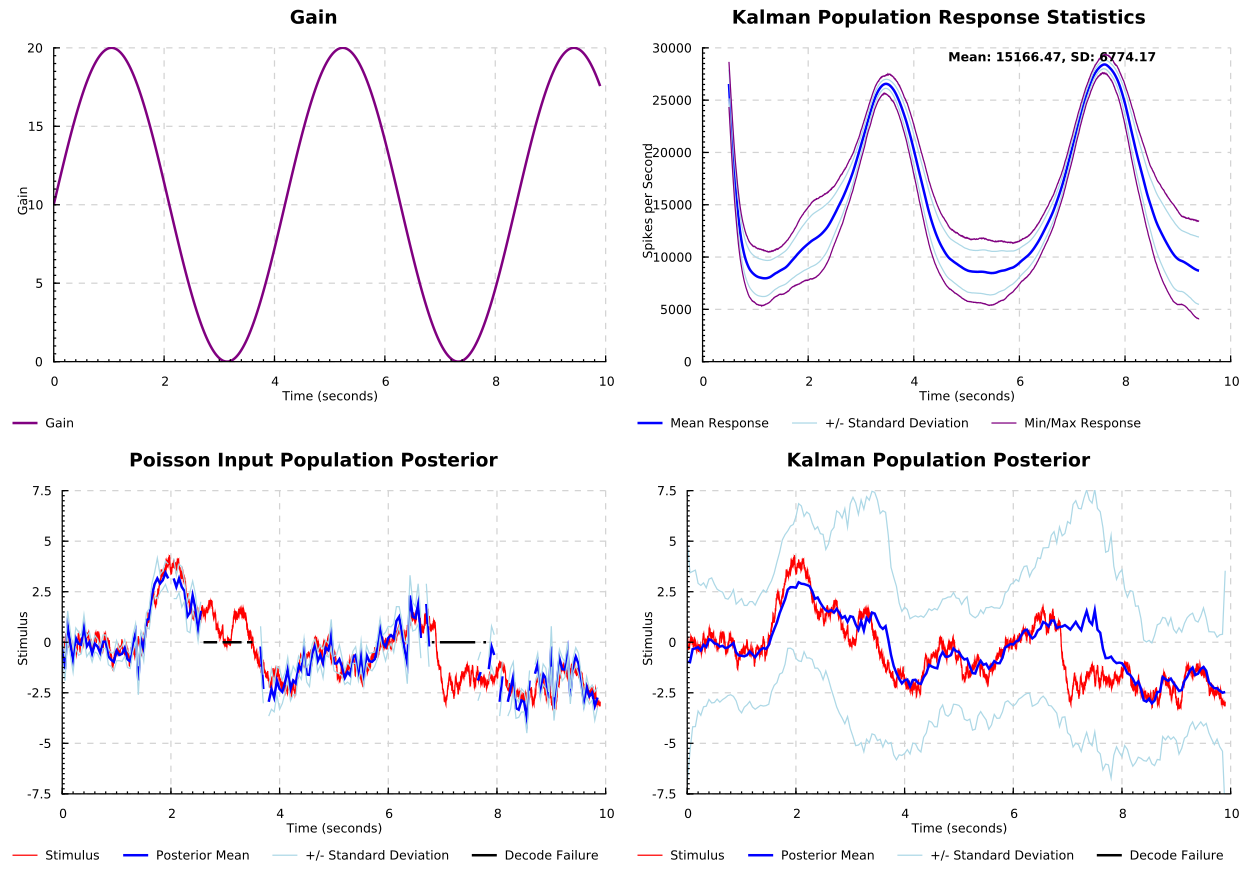}
    \caption{\footnotesize{\textbf{Kalman PPC Simulation}: Top Left) The input population gain. In a vision task, low gain can be interpreted as low contrast which causes the object to become difficult to see. Bottom Left) Decoding the input population over time. When the gain is low, the input population response cannot reliably encode the input. This model is equivalent to the model in (figure~\ref{fig:static1}) replicated over time. Top Right) Spiking statistics of the Kalman population. Bottom Right) Decoding the Kalman population over time. The Kalman population provides a good estimate of the input signal and can interpolate values when the input population isn't active.}}
    \label{fig:kalman}
\end{figure*}

Kalman filters are a class of powerful algorithms for combining observations and model predictions in an optimal way. A 1-dimensional linear Kalman filter is the simplest form of such a filter, but is still capable of solving interesting problems. Where $x$ represents the state and $a$ the model rate, $u$ represents the control signal and $b$ the control rate, and $\eta$ represents the noise, the equation for underlying dynamics of such a Kalman filter has the form of (\ref{eq:kalman}).

\begin{equation}
    \frac{dx(t)}{dt} = a*x(t) + b*u(t) + \eta(t)
    \label{eq:kalman}
\end{equation}

Given the parameters of this equation $a$, $b$, and $\eta$, a control signal $u(t)$, and an estimate of the current state $x(t)$, the Kalman filter will make an \textit{a priori} estimate of $x(t+\Delta t)$ based on the model dynamics. Then, given a noisy observation $z(t+\Delta t)$, the Kalman filter will combine this with the a priori estimate to produce an \textit{a posteriori} estimate which optimally combines the two signals weighted by their variances. Where $a$ is equal to some diffusion parameter $\lambda$ and $b$ is $0$, equation (\ref{eq:kalman}) models a simple diffusion process. Given this model and a series of observations, a Kalman filter can infer the time evolution of this process.

The algorithm for calculating a Kalman filter update step involves solving a series of equations, and the posterior distribution of the probabilistic computation which respects these Kalman equations is given by $p(\tilde{\mathbf{x}}(t+\Delta t) \mid \tilde{\mathbf{x}}(t),\tilde{\mathbf{z}}(t+ \Delta t))$. It is possible for a PPC to implement this probabilistic computation, satisfying an optimality constraint analogous to that given in equation~\ref{eq:optimal}. If we interpret the rates of the Poisson likelihoods from the beginning of this section as rates for a set of Poisson processes, we can model a PPC over time as a set of stochastic differential equations which define the firing rates of the individual neurons of the population (\ref{eq:kalmanppc}). 

\begin{equation}
    \frac{d \tilde{\mathbf{x}}}{d t} = \mathbf{W \cdot \tilde{x} + U \cdot \tilde{x} + M \cdot \tilde{z} - \tilde{x} \cdot Q \cdot \tilde{x}}
    \label{eq:kalmanppc}
\end{equation}

In this equation, the encoding $\tilde{\mathbf{x}}$ of the posterior distribution over the stimulus $p(s \mid \tilde{\mathbf{x}})$ evolves over time. In accordance with equation~\ref{eq:kalman}, $\mathbf{W}$ implements the model rate, $\mathbf{U}$ implements the control rate, and $\mathbf{M}$ implements the contribution of observation.

The purpose of $\mathbf{Q}$ is to act as a regularizer on the time evolution of the firing rates, suppressing firing rates when they are too high. The $\mathbf{Q}$ term in equation~\ref{eq:kalmanppc} is negative and quadratic in the population activity $\tilde{\mathbf{x}}$. In simulating the set of differential equations~\ref{eq:kalmanppc}, the $\mathbf{Q}$ term scales the activity in proportion to a quadratic function of the activity, thus implementing a form of divisive normalization. The results from my simulation of a Kalman PPC on a simple diffusion process are given in figure~\ref{fig:kalman}.

A few interesting observations can be made of the results of this simulation. In figure~\ref{fig:kalman}, the sinusoidal gain depicted in the top left plot results in distinct yet related effects in each other plot. When the gain is low, the input population fires too rarely to reliably encode the stimulus, and attempting to decode a posterior in these cases results in numerical errors (Bottom Left). Nevertheless, even when few spikes are being transmitted from the input population to the Kalman population, the Kalman population is able to interpolate the missing values (Bottom Right). At the same time, a lack of observations causes the standard deviation of the posterior distribution to increase, indicating a declining confidence in the estimate of the state of the system. Interestingly, the population response statistics of the Kalman population show that regions of low confidence correspond to higher firing rates (Top Right). However, as the firing rates of the individual neurons decrease, the range of firing rates exhibited by the population increases. Thus, the Kalman population improves its confidence not by increasing its mean firing rate, but by maximizing the differences amongst the activities of its population.

The top right plot also indicates that the neurons in the Kalman population exhibit exceptionally high firing rates. These firing rates are not biologically realistic, but are required in order to achieve a high enough signal to noise ratio. This could be more realistically modelled by simulating a much larger population of neurons, but simulating a realistic number of neurons is difficult on a home computer since the calculations depend on matrices which scale in size to the square of the number of neurons. Nevertheless, it's not unreasonable to interpret each neuron in the simulation as representing an ensemble of neurons, thereby supporting a more realistic interpretation.

Another problem with the model as implemented is its relatively inefficient use of spike information. Spike information efficiency could be optimized based on other computational approaches \citep{deneve_bayesian_2008,hennequin_stdp_2010}, providing another way to improve the signal to noise ratio and allowing us to lower the firing rate of the PPC.

\section{Recent Empirical Findings}
\label{sec:empirical}

The goal of this paper is to present a biologically realistic model of eye tracking of targets within the PPC framework. The structure of the model has been developed by surveying the current literature and attempting to navigate a path of least controversy through contemporary debates. The primary issues of debate within the literature concern the relative strengths of eye position signals deriving from proprioception versus those deriving from efference copies, and role of gain field modulation in neural computation. In this section I will review some results from these debates which have shaped my model.

\subsection{Efference Copies vs Proprioception}
\label{sec:efference}

In order to implement coordinate transformations, the parietal cortex requires information about eye position. The two primary candidates for being sources of this information are proprioception and efference copies. Proprioception is the sense of the relative positions of various parts of the body, and the signals it produces derive from various sources which monitor the states of muscles, tendons, and skin. Efference copies are corollary discharges produced by the transmission of motor signals. Efference copies are usually construed as copies of the motor commands themselves, and are only transformed into signals pertaining to body positions via a forward model of the effect of the motor command.

Proprioception has the capacity to produce static (as opposed to differential) information about eye position. Nevertheless, it is often assumed to play a role only in the fine tuning of an efference copy based system. The primary reason for this is the apparent delay from the time of motor action to the proprioceptive feedback. For signals measured in the somatosensory cortex, contemporary research indicates the delay to be between 60 and 100 ms \citep{wang_proprioceptive_2007,xu_time_2011}. Nevertheless, saccades are usually made only about 3 times per second, and between each saccade the eye tends to remain still. Thus, somewhere from 60 to 100 ms after a saccade, the delayed proprioceptive information remains still and becomes shift invariant, and can therefore be treated as accurate and contemporary.

Since motor commands are relative to the current state of the system being controlled, efference copies provide a differential representation of state. After being fed through the forward model, this representation can be integrated to form a static representation of eye position. To the extent that this signal is fast, reliable, and the forward model can be well tuned, it is rarely argued that efference copies do not play some important role in providing eye position information. The primary debate rather revolves around whether the proprioceptive signal is combined online with the efference copy to form an estimate of eye position, or whether the efference copy is used to derive the online estimate, with proprioception playing only a post estimate feedback role \citep{ziesche_computational_2011,balslev_eye_2012,wang_eye_2012}.

In spite of some desirable properties, the differential, forward model based position estimate provided by the efference copy system suffers from two important draw backs. Firstly, some argue that rather than there needing to be a distinct integrator of the differential efference copy representation, the differential information could be integrated directly into variable encodings themselves. The problem with this approach is that in dealing with discontinuous stimuli such as blinking targets, when the representation of a stimulus temporarily breaks down, the constant of integration of the eye position is lost as well. Thus, after a blink the differential information is no longer sufficient until a resetting of the constant of integration takes place. Secondly, even where we independently integrate the differential signal, without feedback the standard deviation of a purely forward model based approach eventually diverges.

\subsection{Precision vs Gain Field Modulation}

As described earlier, we have two uses for the gain of a PPC - we can either use it to encode precision information about the encoded posterior, or use it to encode information about a second variable of interest. In the first case, we rely on divisive normalization in order to optimize the precision of the encodings of downstream populations. In the second case, we discard the precision information and effectively turn a PPC which computes a Gaussian posterior into a 2-dimensional encoder.

Previously I had characterized the relationship between the choice of precision versus gain field modulation as something which is subject to debate in the literature, but this is not actually the case. Rather, the literatures studying optimal encoding and gain field modulation are distinct, and when developing a neural circuit there is little information about the relative merits of these two encoding regimes for deciding which regime is appropriate.

As \cite{chang_using_2009} and \cite{blohm_simulating_2012} show, there are different ways to use gain field modulation to accomplish complex transformations. A key characteristic which both of these studies share is that the targets which they are modelling compute transformations the outputs of which are given to the motor system. That is, the outputs of the studied neural circuits occur where there is the need to compute an actual estimate of the stimulus. In the case of Gaussian posteriors, if we use something such as the maximum a posteriori (MAP) estimate to evaluate our stimulus, the estimate of the stimulus will simply be the mean. Thus, the variance of the encoded Gaussian is free to be used for some other aspect of the probabilistic computation.

However, when the response of a given neural circuit is likely to be synthesized with other responses downstream, it is in most cases necessary to retain precision information in order to sensibly weigh the inputs and thus compute the optimal output of the neural circuit. This dichotomy is reflected in the literature as well, where models implementing gain field modulation tend to be of circuits in the parietal cortex, and those implementing divisive normalization tend to be of low level sensory systems.

\section{Model}

In Bayesian statistics, a recursive Bayesian estimator - of which the Kalman filter is a simple form - is an algorithm for combining a forward estimate with an observation. Where the efference copy system drives the control rate matrix, and proprioception provides our observations, the problem delineated in section~\ref{sec:efference} of estimating eye position with proprioceptive and efference copies can be formulated as a problem of recursive Bayesian estimation, and potentially solved by the neural circuit defined in section~\ref{sec:kalman}.

In translating the delineated problem into one which can be solved by a Kalman PPC, one still needs to address the question of what to do about the delayed proprioceptive signal, as a Kalman filter generally assumes that the observation corresponds to the same time step as the forward prediction. 

A low gain results in a low precision observation, which causes a Kalman filter to rely on its forward prediction. As explained previously, the delayed proprioceptive signal is accurate from at least 100ms after a motor command. Thus, if the efference copy system can also be used to inhibit the response the proprioceptive signal for 100ms after any non zero motor command, the proprioceptive signal will be inhibited when it is delayed, used when it is effectively contemporary, and thus combined with the forward prediction at exactly those times when it is effective to do so.

\begin{figure}
    \centering
    \includegraphics[width=0.45\textwidth,clip=true]{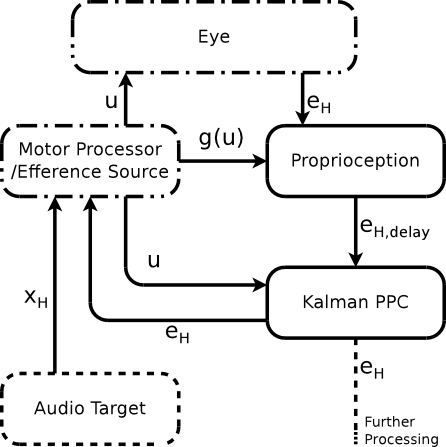}
    \caption{\footnotesize{\textbf{Eye Control Model}: Representation of the eye control model. The dashed boxes represent signal sources, the solid boxes represent PPCs, and the semi dashed boxes represent systems which are neither simple sources nor PPCs. Target and eye position signals are represented by $x$ and $e$, control and gain signals by $u$ and $g$, and head centred coordinate systems are indicated by the subscript and ${H}$.}}
    \label{fig:eyecontrol}
\end{figure}

Omitting the tildes indicating encodings, the neural circuit which I have simulated is depicted in figure~\ref{fig:eyecontrol}. We begin with the movement of the eye caused by the motor command $u(t)$.  Where $d$ represents the length of the proprioception response delay, eye movement causes proprioception to transmit a delayed posterior over eye position in head centred coordinates $p(e_{H}(t-d))$ modulated by the gain $g(u(t))$ transmitted to it from the motor processor/efference copy source. The Kalman PPC integrates this gain modulated signal with the forward model driven by $u(t)$ and the PPC's internal activity. The Kalman PPC then passes its estimate of the contemporary posterior $p(e_{H}(t))$ to the motor processor. Given some target information in head centred coordinates $x_{H}$ (e.g. audio target information), the motor processor combines $x_{H}$ with the MAP of $p(e_{H}(t))$ to produce a new control signal $u(t)$.

The dashed output from the Kalman PPC points towards other uses of the eye position estimate. The phenomenon of gain modulation has so far been observed primarily as supporting the computation of complex coordinate transformations. Since no coordinate transformations are in fact computed in my simple model, sophisticated gain modulated computation of this kind is not required. Nevertheless, $e_{H}$ is an important signal often encoded via gain field modulation, e.g. as the variance of a posterior over some target position in retinal coordinates $x_{R}$, and this output signal could be used effect this gain modulation. In this vein, a more general version of the PPC reach model implemented in \cite{beck_marginalization_2011} could be driven by the signal $e_{H}$. Through this model gain field modulation could be explored within the PPC framework, by observing how $e_{H}$ affects/modulates downstream populations.

The simple saccadic eye control problem solved by my model evolves in the following way. Every 0.3 seconds a new stimulus position is drawn uniformly from the set $\{-2,-1,0,1,2\}$. The goal of the model is to track the target, and so given a target position and an estimate of eye position, the motor processor moves the eye at its maximum speed to minimize the difference between the target and eye positions. In order to suppress oscillatory feedback loops, a motor command is not generated when the difference calculated is smaller than some $\epsilon$.

The Kalman filter is a recursive algorithm, and so in order to ensure good performance, it is important that the internal network dynamics of a Kalman PPC be initialized with a stable and precise estimate of the contemporary system state. As such, simulations are begun with a 2 second initialization period where the stimulus, eye position, and control signal are set to $0$ in order to allow the Kalman PPC to accumulate evidence for its internal model, thus allowing transients to expire before we apply the Kalman PPC to problem in question.

\section{Results}

\begin{figure}
    \centering
    \includegraphics[width=0.475\textwidth,clip=true]{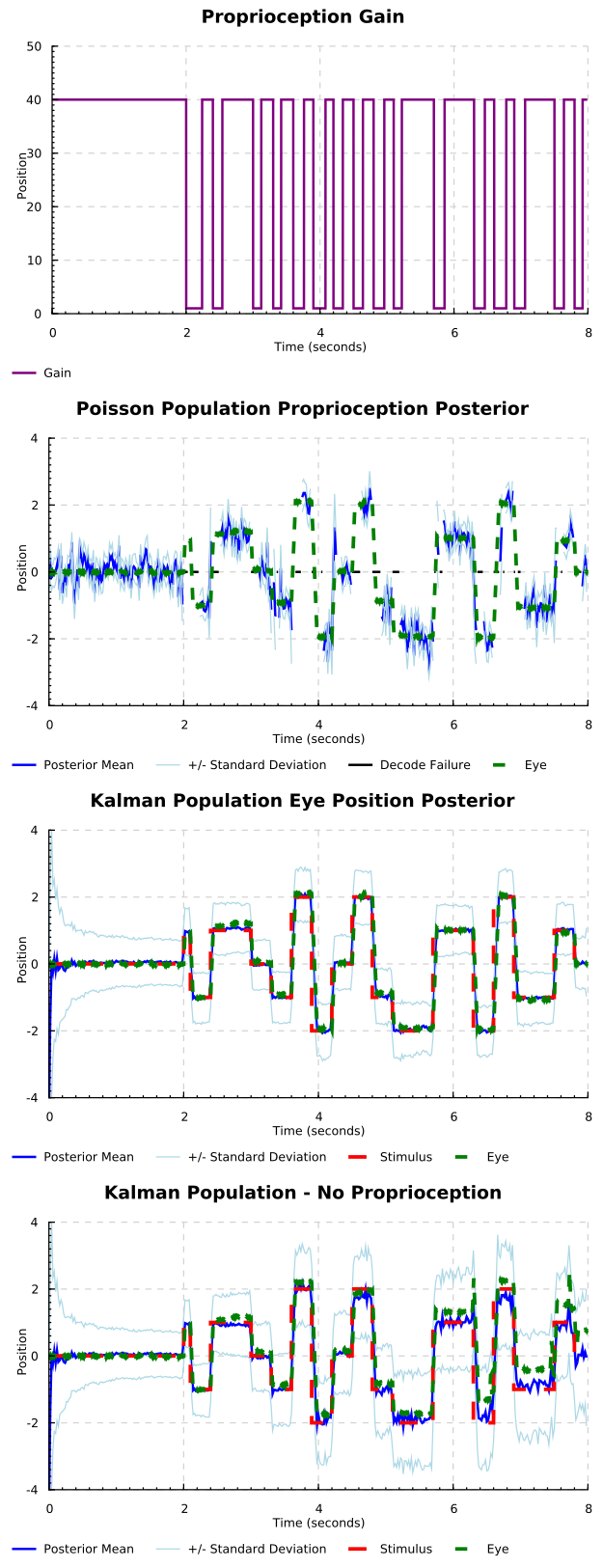}
    \caption{\footnotesize{\textbf{Eye Control Simulation}: 1) The gain used to scale the activity of the proprioception PPC. 2) The decoding over time of the proprioception PPC resulting from a simulation of the model. 3) The decoding over time of the Kalman population. The Kalman population successfully infers the position of the eye and drives the solution to the control problem. 4) In different case in which after initialization, the proprioceptive signal is withheld. The decoding eventually diverges.}}
    \label{fig:circuit}
\end{figure}

The results of my simulations of the model presented in the previous section are plotted in figure \ref{fig:circuit}.

The first plot depicts the gain used to scale the activity of the PPC encoding the eye proprioception signal. The time evolution of the gain begins with the initialization period where it remains at its maximum value. After initialization, whenever a non zero control signal is sent to the motor processor, the gain of the proprioceptive PPC is set to near $0$ (a gain of $0$ is not possible as Poisson distributions require non zero firing rates). This implements the previously explained gain field modulation gating mechanism.

The second plot depicts the simulated proprioception PPC as driven by the gain in the first plot and the state of the eye. Close inspection reveals that after every jump in eye position, a range exists where the eye position cannot be decoded from the proprioception PPC. The purpose of this gain modulation is to filter out those parts of the proprioceptive encoding where the delayed and contemporary signals differ. Thus, all that remains in the activity of the PPC is an encoding of the contemporary position.

The third plot depicts the successful inference of eye position, and the successful tracking of the target stimulus with the eye. The initialization period drives the standard deviation of the encoding down to a minimum, providing the PPC with enough information to make accurate forward predictions. After the initialization we have only intermittent access to the proprioceptive signal. The regular dropping out of the proprioceptive signal results in little effect on the quality of the estimate, and the estimate is accurate enough to drive the motor processor to successfully track the target stimulus.

Corroborating the claim made in section~\ref{sec:efference}, the final plot of figure \ref{fig:circuit} displays a modified version of the complete simulation. In this case we initialize the network as per usual, but then set the gain of the proprioception PPC nearly to zero for the rest of the simulation. The result, as expected, is the slow divergence of the prediction, and eventually divergence of eye control itself. By the end of the simulation the tracking of the target has almost completely decohered.

\section{Discussion}

In summary, in this paper I have presented a PPC neural circuit which efficiently solves the saccadic eye control problem while satisfying certain mathematical and biological constraints. Mathematically, my model optimizes the combining of proprioceptive signals with efference copy signals through a combination of a Kalman filter with a gain modulation based gating mechanism. Biologically the model fits well the observations made in section~\ref{sec:empirical}, and theoretically the model is implemented as a set of Poisson processes implementing Bayes optimal computations, setting it within established literature on modelling neural activity by Poisson processes, and the Bayesian brain hypothesis.

Further work pointed at by this paper include the simulation of higher dimensional models which rely on nonlinear coordinate transformations. In the simple model which I simulated, no coordinate transformations needed to be computed, and thus gain field modulation was not required to compute solutions to, e.g. higher dimensional non linear coordinate transformations. Nevertheless, it is clear that this is due only to the simplicity of the model, and that a more complex model would require that we investigate computations which would likely necessitate gain field modulation. At the same time, groundwork has been laid for exploring various kinds of gain modulated computation using probabilistic population codes, indicating that this question of the computational necessity of gain field modulation could be well investigated within the PPC framework.

The use of gain as a gated signal provides another way we might look at gain field modulation and the kinds of computation it may help implement. Moreover, findings in \cite{xu_time_2011} and unpublished findings from Xu, Karachi, and Goldberg indicate that immediately after a motor command but prior to the onset of an accurate proprioceptive signal, proprioceptive representations of eye position in the somatosensory cortex are somehow disrupted. This preliminary evidence provides support for the biological plausibility of the gating mechanism implemented here, in particular for the proposal that there might be an interaction between corollary discharges and proprioception. I hope that further investigations into these preliminary experimental findings might be assisted by the results of this paper, towards the goal of bringing clarity to our understanding of neural computation in the parietal cortex.

\bibliographystyle{chicago}
\bibliography{/home/alex404/documents/academia/zotero/library}

\end{document}